\documentclass{siamltex}
\usepackage{lineno,hyperref}
\usepackage{multirow}
\usepackage[english]{babel}
\usepackage{amsmath}
\usepackage{bm}
\usepackage[matha,mathx]{mathabx}
\usepackage{mathrsfs}
\usepackage{amssymb}
\usepackage{algpseudocode} 
\usepackage{algorithm}
\usepackage{graphicx}
\usepackage{amsmath}
\usepackage{color} 
\usepackage{caption}
\usepackage{enumitem} 

\makeatletter
\newcommand*{\rom}[1]{\expandafter\@slowromancap\romannumeral #1@}
\makeatother

\usepackage{here}
\usepackage{graphicx}   

\DeclareMathAlphabet{\mathpzc}{OT1}{pzc}{m}{it}
\title{A low-rank non-convex norm method for multiview graph clustering}

\author{  A. Zahir\thanks{The UM6P Vanguard Center, Mohammed VI Polytechnic University, Green 
City, Morocco.}
\and K. Jbilou\footnotemark[1] \thanks{Université du Littoral Cote d'Opale, LMPA, 50 rue F. Buisson, 62228 Calais-Cedex, France.}
\and A. Ratnani\footnotemark[1]}

\begin{document}
\maketitle

\begin{abstract}
This study introduces a novel technique for multi-view clustering known as the "Consensus Graph-Based Multi-View Clustering Method Using Low-Rank Non-Convex Norm" (CGMVC-NC).
Multi-view clustering is a challenging task in machine learning as it requires the integration of information from multiple data sources or views to cluster data points accurately. The suggested approach makes use of the structural characteristics of multi-view data tensors, introducing a non-convex tensor norm to identify correlations between these views. In contrast to conventional methods, this approach demonstrates superior clustering accuracy across several benchmark datasets. Despite the non-convex nature of the tensor norm used, the proposed method remains amenable to efficient optimization using existing algorithms. The approach provides a valuable tool for multi-view data analysis and has the potential to enhance our understanding of complex systems in various fields. Further research can explore the application of this method to other types of data and extend it to other machine-learning tasks.
\end{abstract}

\section{Introduction}
Multiview data clustering is a challenging machine learning task that requires integration data from several data sources or perspectives to accurately group data points. Multiview data naturally arise in various domains, including computer vision, neurobiology, and social network research. The underlying relationships between viewpoints may not be captured well by conventional clustering approaches that examine each view independently, yielding less-than-ideal clustering outcomes. Consequently, developing techniques that effectively leverage information across multiple views is crucial.\\
One widely used approach for multiview data analysis is tensor decomposition\cite{Kolda2009}. By representing multiview data as high-order tensors, tensor decomposition extracts latent components that capture the relationships between views. Popular techniques, such as non-negative matrix factorization (NMF) and non-negative tensor factorization (NTF), have been successfully applied to multiview clustering \cite{Acar2009}.\\
In this paper, we propose a new method for multiview clustering that utilizes a non-convex tensor norm. The proposed method CGMVC-NC utilizes the tensor structure of multiview data and introduces a non-convex tensor norm to capture the correlations between views. The non-convexity of the tensor norm allows it to model complex and non-linear relationships between views. Compared to traditional methods, the proposed method achieves high clustering accuracy on several benchmark datasets.

\noindent 
The contributions are summarized as follows:
 \begin{enumerate}
 \item We propose a novel multiview clustering method that performs spectral embedding and tensor representation with a parametric non-convex tensor norm.
 \item We propose an alternating optimization algorithm to solve the proposed problem. The experimental results validated our approach.
 \end{enumerate}
 
The remainder of this paper is organized as follows. In section 2, we will give some preliminaries to include the tensor related theories as norm and multiplication. In Section 3, we review related work on multi-view clustering and tensor decomposition. In Section 4, we present the proposed method for multi-view clustering using a non-convex tensor norm. In Section 5, we evaluate the proposed method on several benchmark datasets and compare it to state of art methods. Finally, in the final section, we conclude the paper and discuss future research directions.

\section{Preliminaries}
This section presents some preliminaries and notations related to this subject. We focus on third-order tensors $\mathcal{X} \in \mathbb{R}^{N_{1} \times N_{2} \times N_{3}}$.
We define the frontal slices of $\mathcal{X}$, denoted by $X^{(1)},\ldots, X^{\left(N_{3}\right)}$, referred to fixing the last index.\\
We define $\widetilde{\mathcal{X}}$ as the discrete Fast Fourier transform (FFT) of $\mathcal{X}$ along the third dimension, that is, in MATLAB syntax, $\widetilde{\mathcal{X}}=\operatorname{fft}(\mathcal{X},[\,], 3)$.\\
The transpose of $\mathcal{X}$ denoted by $\mathcal{X}^\top \in \mathbb{R}^{N_{2} \times N_{1} \times N_{3}}$ is obtained by transposing each frontal slice and then reversing the order of transposed frontal slices 2 through $N_{3}$.\\
A tensor is called, orthogonal if it satisfies the relationship:
$$\mathcal{X}^\top * \mathcal{X}=\mathcal{X} * \mathcal{X}^\top=\mathcal{I}.$$
Where $*$ referred here \cite{Kilmer2011} to the t-product. 
A tensor is called, f-diagonal if each of its frontal slices is a diagonal matrix.
The T-SVD decomposition theorem \cite{Kilmer2011,Kilmer2013,Hao2013} states that $\mathcal{X}$ can be factorized as: 
\begin{equation}
\mathcal{X}=\mathcal{U} * \mathcal{S} * \mathcal{V}^{*}, 
\end{equation}
where $\mathcal{U} \in \mathbb{R}^{N_{1} \times N_{1} \times N_{3}}, \mathcal{V} \in \mathbb{R}^{N_{2} \times N_{2} \times N_{3}}$ are orthogonal and $\mathcal{S} \in \mathbb{R}^{N_{1} \times N_{2} \times N_{3}}$ is f-diagonal tensor. Next, we define some norms, starting with the Frobenius norm and the nuclear norm\cite{Lu_2020}:
\begin{equation}
\begin{aligned}
\|\mathcal{X}\|_{F}^{2}&=\sum_{i,j,k}\mathcal{X}_{i j k}^2.\\
\|\mathcal{X}\|_{\circledast}&=\sum_{i=1}^{N_{3}}\left\|\widetilde{X}^{(i)}\right\|_{*},
\end{aligned}
\end{equation}
where $\left\|.\right\|_{*}$ is the nuclear norm of matrix, which is exactly the sum of the singular values. It is proven \cite{Lu_2020} to be a valid norm and the tightest convex relaxation to the norm of the tensor average rank within the unit ball of the tensor spectral norm. \\
We can further define a generalization of this norm called the weighted tensor nuclear norm :
$$
\|\mathcal{X}\|_{\omega, \circledast}=\sum_{i=1}^{N_3}\left\|\widetilde{X}^{(i)}\right\|_{\omega, *}=\sum_{i=1}^{N_3} \sum_{j=1}^{\min \left(N_{1}, N_{2}\right)} \omega_{j} \sigma_{j}\left(\widetilde{X}^{(i)}\right) \text {, }
$$
where $\sigma_{j}(Z)$ denotes the $j-$largest singular value of the matrix $Z$.\\
Another norm that will be presented is the t-Gamma tensor quasi-norms presented in \cite{Cai2019}, defined as:
$$
\|\mathcal{X}\|_{t-\gamma}=\frac{1}{N_3} \sum_{n_3=1}^{N_3}\left\|\widetilde{X}^{(n_3)}\right\|_\gamma,
$$
where, $\|.\|_\gamma$ represents the Gamma quasi-norm for matrices \cite{Kang_2015} , for $Z \in \mathbb{R}^{N_1 \times N_2}$
$$\|Z\|_\gamma=\sum_{i=1}^{\min(N_1, N_2)} \frac{(1+\gamma) \sigma_i(Z)}{\gamma+\sigma_i(Z)}, \gamma>0.$$

\section{Related work}
Multiview clustering \cite{zahir2023Review} approaches can be roughly divided into two types based on how the similarity graph is constructed: subspace segmentation-based methods and graph-based methods.\\
Multi-view subspace clustering attempts to discover a latent space from all the different views to deal with high-dimensional data and then applies a traditional clustering approach to this subspace.\\
Low-Rank Representation LRR \cite{Liu_2010} and Sparse Subspace Clustering SSC \cite{Ehsan} are two widely used single-view subspace clustering techniques. Building on these methods, several multiview subspace clustering approaches have been developed \cite{Luo2018,Cao_2015}. Among them, the Coupled Low-Rank and Sparse Representation CLRS model, proposed in \cite{Wang2013}, integrates both low-rank and sparsity constraints. Experimental results in \cite{Wang2013} show that LRR and SSC misidentify different data points, highlighting their complementary nature. Therefore, coupling low-rank and sparsity criteria is crucial for effectively capturing both the global and local structures of the data.\\
A different approach \cite{liu2022stationary}, considers regarding graphs as Markov chain, with a co-supervising strategy with structure information. 
Treating data as matrices rather than as a full tensor implies that the relationship between subspaces is overlooked; thus, various approaches \cite{Wu_2018,XIE202157,Cheng2019} transform data into a third-order tensor to capture high-order view correlations.\\
Multi-view graph clustering aims to find a fusion graph of all views, then applies the classical clustering methods on the fusion graph, as spectral clustering or graph-cut algorithms to produces the final clustering. In
\cite{Chaudhuri2009}, canonical correlation analysis CCA is used to identify the common underlying structure between different views. Although it does not give importance to the weight. \cite{Xia2010} adds new hyper parameters. \cite{Nie_2016} presented a method named AMGL without introducing any hyper parameters. AMGL can learn the weight by itself. In \cite{Nie_2017, Jing2017}, these methods are based on the generalization of the Laplacian rank constrained to give a similarity matrix with exactly the desired number of components. As in the subspace based method \cite{Chen2019,Haiyan2021,Zhao2022,li2021consensus}, data is transformed into a tensor to capture the high-order view correlations within the data.\\
Noise and redundant information are typically mixed with the original characteristics in practice. Consequently, the learned consensus similarity graph may be incorrect. To overcome this problem, \cite{li2021consensus}, proposed an approach called Consensus graph learning CGL, which trains an adaptive neighbor graph in a new low-dimensional embedding space rather than the original feature space.\\
CGL method composed of \textbf{three steps}: First, it learns the view-specific similarity graph $S^{(v)}$ using an adaptive neighbor graph learning:
\begin{equation}
\begin{aligned}
\forall i=1,...,n \quad &\min_{S^{(v)}_{i}} \sum_{j=1}^{n}\left\|\mathbf{x}_{i}^{(v)}- \mathbf{x}_{j}^{(v)}\right\|_{2}^{2} S^{(v)}_{i j} + \gamma s_{i j}\\
& s^{(v)}_{i j} \geq 0, \mathbf{1}^\top \mathbf{s}^{(v)}_{i}=1, j=1, \ldots, n,
\end{aligned}
\label{eq: CGL_step1}
\end{equation}
the new low-embedding space is learned as follows:
\begin{equation}
\begin{aligned}
\min_{F^{(v)}, \mathcal{T}} & -\lambda \sum_{v=1}^{V} \operatorname{tr}\left(F^{(v)} A^{(v)} F^{(v)^\top}\right)
+\frac{1}{2}\|\mathcal{F}-\mathcal{T}\|_{F}^{2}+ \|\mathcal{T}\|_{w,*}\\
\text { s.t. } & F^{(v)^\top} F^{(v)}=I_c, v=1,\ldots,V \\
&\mathcal{F}=[\overline{F}^{(1)} \overline{F}^{(1)^\top};\cdots;\overline{F}^{(n)} \overline{F}^{(n)^\top}],
\end{aligned}
\label{eq: CGL_step2}
\end{equation}
where the normalized spectral embedding matrix $\overline{F}^{(v)}$ is obtained by normalizing the rows of $F^{(v)}$, the normalized affinity matrix is given
\begin{equation}
A^{(v)}=D^{(v)^{-0.5}}S^{(v)}D^{(v)^{0.5}},
\label{eq:A}
\end{equation}
where $D^{(v)}$ is a diagonal matrix containing the sum of elements of the affinity matrix and the tensors $\mathcal{F}, \mathcal{T} \in \mathbb{R}^{n \times V \times n}$, to which the consensus graph $S$ can be learned in the last step using the adaptive neighbour again as:
\begin{equation}
\begin{aligned}
\min_{S} & \sum_{v=1}^{V} \left(\sum_{i, j=1}^{n}\left\| \overline{\textbf{f}}_{i}^{(v)} - \overline{\textbf{f}}_{j}^{(v)} \right\|_{2}^{2} s_{i j} + \frac{\gamma}{V} s_{i j} \right)  \\
& s_{i j} \geq 0, \mathbf{1}^\top \mathbf{s}_{i}=1, i, j=1, \ldots, n.
\end{aligned}
\label{eq: CGL_S}
\end{equation}

\section{Proposed method}
\subsection{Data}
Given multi-view data $\{ X^{(v)} \}_{v=1,\dots,V}$ where $X^{(v)}=[\textbf{x}_{1}^{(v)}, \ldots, \textbf{x}_{n}^{(v)}] \in \mathbb{R}^{d_v \times n}$ is the data matrix of the $v$-th view, $d_v$ is the dimension of features in view $v$, $n$ is the number of data points and $V$ is the number of views. The goal of multi-view clustering is to learn a consensus clustering result from multiple views. In this paper, we assume that the data points are from $k$ clusters, and the clustering result $Y \in \mathbb{R}^n$ is a discrete vector with $Y_i \in \{1,\dots,k\}$, $i=1,\dots,n$. The clustering result $Y$ is unknown and needs to be learned from the multi-view data.
\subsection{Graph construction}
The graph constrction is the first step in the method, the most common method are the k-nearest neighbor (kNN), the $\epsilon$-neighborhood graph and the complete graph. The kNN graph is constructed by connecting each data point to its k-nearest neighbors. The $\epsilon$-neighborhood graph is constructed by connecting each data point to its neighbors within a radius $\epsilon$. The complete graph put an edge between all points. The result is an affinity matrix, where the weight of the edge, is commonly computed using:
\begin{itemize}
\item Gaussian kernel: $s_{i j}^{(v)}=\exp \left(-\frac{\left\|\mathbf{x}_{i}^{(v)}- \mathbf{x}_{j}^{(v)}\right\|_{2}^{2}}{2\gamma}\right)$, where $\gamma$ is a scalling parameter.
\item Cosine similarity: $s_{i j}^{(v)}=\frac{\mathbf{x}_{i}^{(v)^\top} \mathbf{x}_{j}^{(v)}}{\left\|\mathbf{x}_{i}^{(v)}\right\|_{2}\left\|\mathbf{x}_{j}^{(v)}\right\|_{2}}$
\item Euclidean distance: $s_{i j}^{(v)}=\left\|\mathbf{x}_{i}^{(v)}- \mathbf{x}_{j}^{(v)}\right\|_{2}$
\item Binary: $s_{i j}^{(v)}=1.$
\end{itemize}
In this paper, we will stick with the adaptive neigbhour graph, which is a combination of the kNN and the $\epsilon$-neighborhood graph. The adaptive neighbor graph is constructed by connecting each data point to its k-nearest neighbors, and then, for each data point, the weight of the edge is computed as the distance between the data point and its k+1 nearest neighbor. The weight of the edge is computed using the Euclidean distance.

\subsection{Problem formulation}
Since in equation \eqref{eq: CGL_step2}, Since we've previously attained the tightest convex low-rank as a norm, we suggest loosening the convexity restriction using the Non-Convex Low Rank to improve the probability of reaching the norm, i.e. the t-Gamma tensor quasi-norm. The Gamma tensor quasi-norm has some algebraic properties. It is positive definite and unitarily invariant, its limits are exactly the tubal rank when $\gamma \to 0$, and the tensor nuclear norm when $\gamma \to +\infty$. It has shown some promising results in terms of accuracy and efficiency \cite{Kang_2015}. It can overcome the imbalanced penalization by different singular values in the convex nuclear norm. Therefore, we propose a new approach called Consensus graph-based multi view clustering using a low-rank non convex norm (CGMVC-NC). We suggest to change problem \eqref{eq: CGL_step2} to become:
\begin{equation}
\begin{aligned}
\min_{F^{(v)}, \mathcal{T}} & \left[ -\lambda \sum_{v=1}^{V} \operatorname{tr}\left(F^{(v)} A^{(v)} F^{(v)^\top}\right)
+\frac{1}{2} \|\mathcal{F}-\mathcal{T}\|_{F}^{2} +\rho \|\mathcal{T}\|_{t-\gamma} \right] \\
&\text { s.t. } F^{(v)} \in V_c(\mathbb{R}^n), v=1, \cdots, V \\
&\mathcal{F}=[\overline{F}^{(1)} \overline{F}^{(1)^\top};\cdots;\overline{F}^{(V)} \overline{F}^{(V)^\top}].
\end{aligned}
\label{eq: Proposed_method}
\end{equation}
where the Stifeld manifold $V_c(\mathbb{R}^n)$ is the set of all $n \times c$ matrices with orthonormal columns.\\
We can provide the framework of the problem as follows:
\begin{figure}[H]
\centering
\includegraphics[scale=.45]{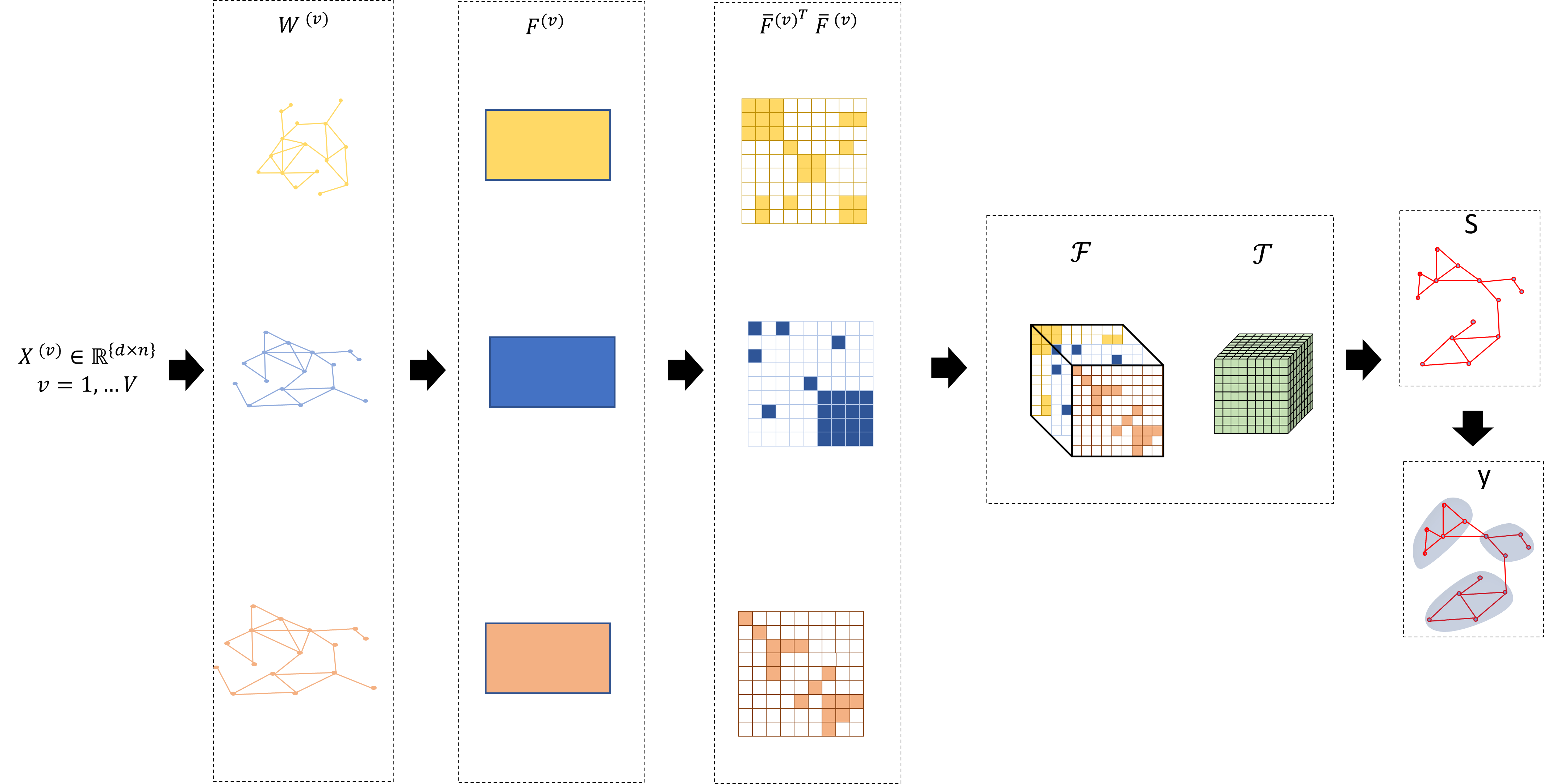}
\caption{The proposed method framework}
\end{figure}

\subsection{Solution}
Since the problem consists of mainly 3 steps, each step can be solved separately.\\
The solution of the \textbf{first} step \eqref{eq: CGL_step1}, which is called, an adaptive neighbor learning, is given in \cite{Nie_2014} and is a closed form solution: 
\begin{equation}\label{sol:first_step}
s^{(v)}_{i j}=\left(-\frac{\left\|\mathbf{x}_{i}^{(v)}- \mathbf{x}_{j}^{(v)}\right\|_{2}^{2}}{2\gamma} + \eta_i  \right)_{+},
\end{equation}
where $\eta_i$ is the Lagrangian multiplier of the Lagrangian problem of \eqref{eq: CGL_step1} associated to the constraint $ \mathbf{1}^\top \mathbf{s}_{i}=1$.\\
The paper made another assumption to learn sparse similarity matrix as it is shown that is robust to noise and outliers, it can be achieved by letting the point to only connect to their k-nearest neighbour, with this assumption, the value of regularization parameter $\gamma$ is set to the meaning of all $\gamma_i$ \cite{Nie_2014} 
\begin{equation}
\gamma=\frac{1}{2n}\sum_{i=1}^n \left(k\left\|\mathbf{x}_{i}^{(v)}- \mathbf{x}_{k+1}^{(v)}\right\|_{2}^{2}-\sum_{j=1}^{k}\left\|\mathbf{x}_{i}^{(v)}- \mathbf{x}_{j}^{(v)}\right\|_{2}^{2} \right)
\end{equation}
and $\left\|\mathbf{x}_{i}^{(v)}- \mathbf{x}_{j}^{(v)}\right\|_{2}^{2}$ are ordered from small to larger with respect to $j$. $\eta_i$ can be found \cite{Nie_2014} as  
\begin{equation}
\eta_i=\frac{1}{k}+ \frac{1}{2k\gamma_i}\sum_{j=1}^{k}\left\|\mathbf{x}_{i}^{(v)}- \mathbf{x}_{j}^{(v)}\right\|_{2}^{2},\; \gamma_i=\frac{k}{2}\left\|\mathbf{x}_{i}^{(v)}- \mathbf{x}_{k+1}^{(v)}\right\|_{2}^{2}-\frac{1}{2}\sum_{j=1}^{k}\left\|\mathbf{x}_{i}^{(v)}- \mathbf{x}_{j}^{(v)}\right\|_{2}^{2}
\end{equation}

The \textbf{second} step \eqref{eq: Proposed_method}, is an optimization problem over two variables, it is difficult to solve it directly, thus, we will use an alternate optimization algorithm. The problem can be transformed into two sub-problems, i.e., $F^{(v)}$ and $\mathcal{T}$ sub-problems, and we alternately optimize each variable while the rest of the other variables are fixed, as follows:

\textbf{$F^{(v)}$-sub-problem}: By fixing the other variable $\mathcal{T}$ and since the matrices $\{F^{(v)},v=1\ldots,V\}$ are independent between them, we can decompose it to $V$ independent problems, the equation \eqref{eq: Proposed_method} can be rewritten in matrix form as : 
\begin{equation}
\begin{aligned}
& \min _{F^{(v)}} \left[-\lambda \operatorname{Tr}\left(F^{(v) \top} A^{(v)} F^{(v)}\right)+\frac{1}{2}\left\|\overline{F}^{(v)} \overline{F}^{(v) \top}-\mathcal{T}(:, v,:)\right\|_F^2 \right] \\
& \text { s.t. } F^{(v) \top} F^{(v)}=I_c.
\end{aligned}
\label{eq: F_subprob}
\end{equation}
The equation \eqref{eq: F_subprob} can be further rewritten as:
$$
\begin{aligned}
\min _{F^{(v)}} & \left[-\lambda \operatorname{Tr}\left(F^{(v) \top} A^{(v)} F^{(v)}\right)+\frac{1}{2} \operatorname{Tr}\left(\overline{F}^{(v)} \overline{F}^{(v) \top} \overline{F}^{(v)} \overline{F}^{(v) \top}\right) \right. \\
& \left. -\frac{1}{2} \operatorname{Tr}\left(\overline{F}^{(v)} \overline{F}^{(v) \top} \left(\mathcal{T}(:, v,:)+\mathcal{T}(:, v,:)^T\right)\right) \right] \\
\text { s.t. } & F^{(v) \top} F^{(v)}=I_c.
\end{aligned}
$$
Let the diagonal matrix $ P^{(v)} \in \mathbb{R}^{n \times n}$ be:
\begin{equation}
P^{(v)} =\operatorname{diag}\left( || f_i^{(v)} || ^{-1} \right) , i=1,\cdots n,
\label{eq:P}
\end{equation}
where the vector $f_i^{(v)}$ represents the i-th row of $F^{(v)}$. Hence, the normalized embedding matrix is given by:
\begin{equation}
\overline{F}^{(v)}=P^{(v)} F^{(v)}.
\label{eq:F_normalized_F}
\end{equation}
The optimization problem can be further rewritten as:
\begin{equation}
\begin{aligned}
&\max _{F^{(v)}} \operatorname{Tr}\left(F^{(v) \top} H^{(v)} F^{(v)}\right) \\
&\text { s.t. } F^{(v) \top} F^{(v)}=I_c,
\end{aligned}
\label{eq: F_sol}
\end{equation}
where $H^{(v)}=\lambda A^{(v)}+\frac{1}{2} P^{(v)}\left(\mathcal{T}(:, v,:)+\mathcal{T}(:, v,:)^\top\right) P^{(v)}-\frac{1}{2} P^{(v)} \overline{F}^{(v)} \overline{F}^{(v) \top} P^{(v)}$.\\
\textbf{Remark}:
The problem is now an optimization problem of trace on a manifold. The matrix $H^{(v)}$ depends on the optimization problem $F^{(v)}$, which makes finding the optimal solution of $F^{(v)}$ thought to solve. Another difficulty, is that the function with respect to $F^{(v)}$ is not smooth (which is due the inverse of the norm and hence, non differentiability on point zero), and thus not smooth. 
To manage to give an approximation, we use the Picard method, and since this is a sub-problem, we propose to consider $H^{(v)}$ as a function of the previous iteration, by this consideration, and since $H^{(v)}$ is symmetric, the solution would be well known, according to the Ky Fan's Theorem \cite{Fan1949}, are the eigenvectors normalized corresponding to the $c$ largest eigenvalues of $H^{(v)}$.\\
For the k parameter of the adaptive neighbor, we can use the same value for all views, or use a different value for each view. We can also use a different value in the last step. The numerical result show that the same value if sufficient for all steps.\\

\textbf{$\mathcal{T}$-sub-problem}: By fixing the other variables, the equation \eqref{eq: Proposed_method} can be written as :
\begin{equation}
\min _{\mathcal{T}} \frac{1}{2}\|\mathcal{F}-\mathcal{T}\|_F^2+ \rho \|\mathcal{T}\|_{t-\gamma}.
\label{eq:T_subProb}
\end{equation}
The tensor $\mathcal{T}^*$ that minimizes equation \eqref{eq:T_subProb} is given in \cite{Cai2019} as follows:
\begin{equation}
\mathcal{T}^*=\mathcal{U} * \Omega * \mathcal{V}^{\mathrm{T}},
\label{eq:T_sol}
\end{equation}
where $\mathcal{F} =\mathcal{U} * \mathcal{S} * \mathcal{V}^{\mathrm{T}}$ is the T-SVD of the tensor, and $\Omega \in \mathbb{R}^{n \times V \times n}$ is f-diagonal tensor.\\
For $1 \leq i \leq min(n,V)=V \text{ (since generally } V \leq n), \quad 1 \leq j \leq n$, the $\widetilde{\Omega}_{i, i, j}$ is the limit point of Fixed Point Iteration:
\begin{equation}
\widetilde{\Omega}(i,i,j)^{(k+1)}=\left(\widetilde{\mathcal{S}}(i, i, j)- \frac{ \rho \gamma(1+\gamma)}{\left( \gamma+\widetilde{\Omega}(i, i, j) ^{(k)} \right)^2}\right)_{+},
\end{equation}
where the fixed point converges to a local minimum after several iterations.\\
While the final solution is not the global optimal (NP-Hard), which is the case in most of this type of problems, the experiments show that the algorithm converges to a solution that produces promising results and even gives better results than the methods where the convergence is proven.\\

\textbf{The last step:} is similar to the first one, note that to avoid the need of further clustering algorithm on the obtained affinity matrix, we will make use of the next Theorem:
\begin{theorem}{\cite{mohar1991laplacian,chung1997spectral}}
The number of connected components in the  non-negative similarity matrix of a graph is the same as the multiplicity of the eigenvalue 0 of its corresponding Laplacian matrix.
\end{theorem}

Then, by adding a constraint of $\operatorname{rank}(L_U)=n-c$ ensure that the obtained affinity matrix has exactly c connected components. However, the problem would be hard to tackle, thus the use of Ky Fan' Theorem; Since, $\sum_{i=1}^c \mathbf{v}_i(L_u)=0$ is equivalent to $\operatorname{rank}(L_U)=n-c$, where $\mathbf{v}_i$ corresponds to the i-th smallest eigenvalue. Thus, we can write the problem as:
\begin{equation}\label{step_3}
\begin{aligned}
\quad &\min_{S,Q}  \sum_{j=1}^{n} \left( \sum_{v=1}^{V} \left\|\overline{\textbf{f}}_{i}^{(v)}- \overline{\textbf{f}}_{j}^{(v)}\right\|_{2}^{2} S^{(v)}_{i j}  + \gamma s_{i j} \right)+ \lambda \operatorname{Tr}(Q^TL_SQ)\\
&  Q^TQ=I, Q \in \mathbb{R}^{n \times c}, \;\mathbf{1}^\top \mathbf{s}_{i}=1,\; s_{i j} \geq 0, \; i,j=1, \ldots, n,
\end{aligned}
\end{equation}
Where $\lambda$ is large enough, to make sure the last term $\operatorname{Tr}(Q^TL_SQ)=0$.\\
The problem can be solved using alternative optimization problem:\\
- When $S$ is fixed, the problem becomes $ min \operatorname{Tr}(Q^TL_SQ),  Q^TQ=I, Q \in \mathbb{R}^{n \times c}$, which is exactly Ky fan's Theorem.\\
- When $Q$ is fixed, the problem is equivalent to:
\begin{equation}
\begin{aligned}
&\min_{s_{i}} || s_i + \frac{1}{2V\gamma}d_{i} ||_2^2 \\
& \mathbf{1}^\top \mathbf{s}_{i}=1,\; s_{i j} \geq 0, \;  j=1, \ldots, n,
\end{aligned} 
\end{equation}
with $d_{i,j}= \lambda  \left\|\mathbf{q}_{i}- \mathbf{q}_{j} \right\|_{2}^{2} + \sum_{v=1}^{V} \left\|\overline{\mathbf{f}}_{i}^{(v)}- \overline{\mathbf{f}}_{j}^{(v)}\right\|_{2}^{2} $,
the solution with k-neighbour can be written in closed form as:
\begin{equation}\label{sol:third_step_S}
s_{i j}=\left(-\frac{d_{i j}}{2\gamma} + \eta_i  \right)_{+},
\end{equation}
where $\eta_i,\gamma$ can be found similarly.
\subsection{Algorithm}
The following algorithm solves the last step in the problem:s
\begin{algorithm}[H]
\caption{Solve \eqref{step_3}}
\hspace*{\algorithmicindent} \textbf{Input:} $\mathcal{F}^{(v})$ (data), k (adaptive neighbor), c (cluster number).\\
\hspace*{\algorithmicindent} \textbf{Output:} Affinity matrix $S$.
\begin{algorithmic}[1]
\State Initialize $S$ by an adaptive neighbor without the rank constraint.
\While{Not converged}
\State Update $Q$.  \Comment{Using Ky Fan's Theorem}
\State update $S$. \Comment{Using Eq. \eqref{sol:third_step_S}}
\EndWhile
\end{algorithmic}
\label{algo_1}
\end{algorithm}
We can initialize $\lambda=\gamma$, then, at each iteration, increase it if the number of connected components found is smaller than c, and decrease it if greater than c. This is helpful in practice, and will accelerate the procedure.

The algorithm proposed to solve this problem is as follows:
\begin{algorithm}[H]
\label{algotihm}
\caption{Proposed method}
\hspace*{\algorithmicindent} \textbf{Input:} $\mathcal{X}$ (data), k (adaptive neighbor), c (cluster number), $\gamma$ (Norm parameter), $\lambda$ (balance parameter). $\rho$ (parameter)\\
\hspace*{\algorithmicindent} \textbf{Output:} Consensus graph $S$ , clustering result $y$
\begin{algorithmic}[1]
\State $H^{(v)}=0$ \Comment{Initialization}
\State Compute $S^{(v)}$. \Comment{Via Eq. \eqref{sol:first_step}}
\State Compute $A^{(v)}$. \Comment{Using Eq. \eqref{eq:A}}
\While{Not converged}
\State Update $F^{(v)}$ \Comment{Solution of Eq. \eqref{eq: F_sol}}.
\State Update $P^{(v)}$ and $\overline{F}^{(v)}$ \Comment{Using Eq. \eqref{eq:P} and \eqref{eq:F_normalized_F} }
\State Compute $\mathcal{F}=[\overline{F}^{(1)} \overline{F}^{(1)^\top}; \cdots;\overline{F}^{(n)} \overline{F}^{(n)^\top}]$.
\State update $\mathcal{T}$. \Comment{Using Eq. \eqref{eq:T_sol}}
\EndWhile
\State Compute $S$ \Comment{Via Algorithm \ref{algo_1}}
\State Return the clustering result $y$. \Comment{Using k-means}
\end{algorithmic}
\end{algorithm}
The iteration (while loop) in the algorithm should stop when a condition criteria (convergence) is satisfied. The relative error in the objective function value will be used. We can also use the maximum number of iteration or running time.\\
The relative error can be calculated using the objective function of the previous and the current iteration
$$\dfrac{\operatorname{Obj}(t)-\operatorname{Obj}(t-1)}{\operatorname{Obj}(t-1)}< \epsilon,$$
with $\epsilon$ is a threshold set generally as 1e-6.

\section{Experiments}
In this section, we do some experiments and compare our method with the state of art methods.

\subsection{Data sets}
We used multiple different real world data sets, that are commonly used in the literature. The majority of image multi-view data sets are derived from initial single-view image data sets.
\begin{itemize}
\item \textit{One-hundred plant species leaves data set leafs (100Leaves)} \footnote{https://archive.ics.uci.edu/ml/data sets/One-hundred+plant+species+leaves+data+set}: It consists of 1600 samples, of 100 different species, 4 views are given: shape descriptor, fine scale margin and texture histogram.
\item \textit{3 sources data set (3Sources)} \footnote{http://elki.dbs.ifi.lmu.de/wiki/data sets/MultiView}: It consists of 169 news, provided by three news companies, including the BBC, Reuters, and The Guardian. Each piece of news was manually labeled with one of six relevant designations.
\item \textit{Handwritten digit data set(HW)} \footnote{http://archive.ics.uci.edu/ml/datasets/Multiple+Features}: It is from the UCI repository. It consists of 2000 samples of handwritten digits, with a 6 views, It has 10 clusters (0-9).
\end{itemize}
The characteristics of these raw feature data sets are given in the table \ref{Tab:data_sets} below, which displays the number of samples $n$, the number of views $V$, the cluster number $c$, and the dimension of features in view $d_v$. More details can be found regarding the dataset using the hyperlinks in the footnotes.
\begin{table}[ht]
\begin{center}
\begin{minipage}{\textwidth}
\caption{Summary of the benchmark data sets used}
\label{Tab:data_sets}
\begin{tabular}{lccccccccc}
\hline Data set & \# samples & \# views & \# clusters & $\mathrm{d}_1$ & $d_2$ & $d_3$ & $d_4$ & $d_5$ & $d_6$ \\
\hline
3Sources & 169 & 3 & 6 & 3560 & 3631 & 3068 & - & - & - \\
100Leaves & 1600 & 3 & 100 & 64 & 64 & 64 & - & - & - \\
HW & 2000 & 6 & 10 & 216 & 76 & 64 & 6 & 240 & 47 \\
\hline
\end{tabular}
\end{minipage}
\end{center}
\end{table}

\subsection{Compared methods}
In order to work with the same data set and the same clustering metrics, a few modifications were made to these data sets prior to evaluating the state-of-the-art methods.
\begin{itemize}
\item \textbf{MVGL} \footnote{https://github.com/kunzhan/MVGL} This method optimizes the joint graph with a rank constraint and treats data as matrices without any tensor related used.
\item \textbf{CGL} \footnote{https://github.com/guanyuezhen/CGL} This method integrates adaptive neighbour and finds the fusion graph in a new subspace using the a convex norm.
\item \textbf{MV-LRSCC} \footnote{https://github.com/mbrbic/Multi-view-LRSSC} This method jointly learns a low-rank representation and sparse subspace clustering across multiple views. It has 4 variants, i.e., pairwise, centroid, parirwise kernel, and centroid kernel.
\item \textbf{AMGL} \footnote{https://github.com/kylejingli/AMGL-IJCAI16} This method learns the wights of each graph automatically and without the need of any parameter.
\end{itemize}

\subsection{Evaluation metrics}
To randomize the tests, each algorithm is executed ten times. Using well-known benchmark data sets, several metrics are used to assess how effectively these multi view clustering methods perform. Multiple measurements favor different features in the clustering task, which motivates us to employ more than one.\\
The accuracy (ACC), the normalized mutual information (NMI), the F1 measure (F-measure), and the adjusted rand index (ARI) are the main ones used; see \cite{zahir2023Review} for more details on these measures
A higher number for each measure reflects better clustering performance.

\subsection{Computation complexity}
We can calculate the computation complexity of our method. The first step of computing the similarity matrices needs $\mathcal{O}(nlog(n)V)$, normalizing it needs $\mathcal{O}(n^2V)$, computing $H^(v)$ needs $\mathcal{O}(n^2)$, computing the c largest eigenvectors of $F^(v)$ is $\mathcal{O}(n^2c)$, and computing $\mathcal{T}$ needs $\mathcal{O}(n^2Vlog(n)+n^2V^2+nVw))$, where $w$ is the maximum number of iterating of the fixed point(usually not very large), the first part represent the FFT and IFFT costs, and the second, represents the SVD cost. Computing $S$ takes the same amount as the first step, it needs $\mathcal{O}(nlog(n)$. Finally, the spectral clustering needs $\mathcal{O}(n^2c)$. In total, the complexity of the proposed model is $\mathcal{O}\left( n^2(c+V)+nlog(n)+tn^2V\left(c+log(n)+V+\dfrac{w}{n}\right) \right)$, where $t$ is the number of iteration.\\
We also compare the time cost of each method on each dataset.
\begin{figure}[H]
\centering  \includegraphics[scale=.7]{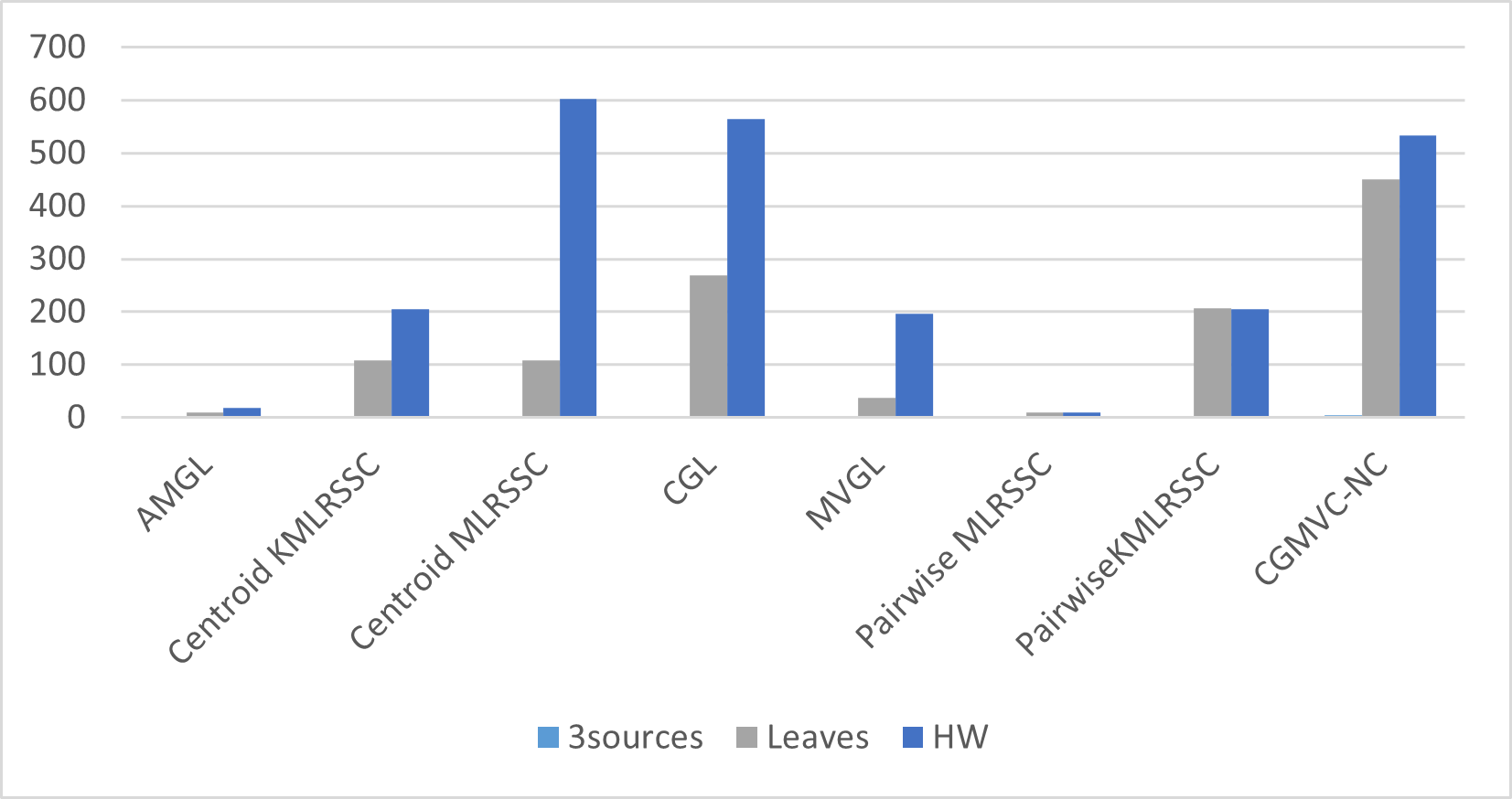}
\caption{Time cost of different methods per second.}
\end{figure}
\noindent
Which shows that our method, even it has a non- convex part, it takes generally as much time as most of the benchmark methods. 
The dataset 3Sources is quite small compared to the others, which is why it is quite has low time cost. 
\subsection{Parameter setting}
The parameters value of each methods were chosen as suggested from their papers, along with other values that are near the suggested ones and we report the best value by accuracy since we have multiple evaluation metrics.\\
In our suggested model, the parameters are $\gamma$ is chosen from the range $ [ 10^{-4} , 10^{-3}, 10^{-2} , \\ 10^{-1} , 1, 10, 10^{2} ] $ , we found that setting $\gamma$ to $0.1$ works best in all cases. For $\lambda$, we follow the suggested in the CGL paper and we set it to $\frac{1}{\sqrt{min(N_1,N_3)}}$. For the parameter $\rho$, we've tried in the$ [ 10^{-4} , 10^{-3}, 10^{-2} , 10^{-1} , 1, 10, 10^{2} ] $, the experiments suggests that the value has not a decisive effect.\\
For some methods, a post clustering (e.g., k-means) is needed on the obtaining embedding data, as it is sensitive to initial values, we repeat the k-means clustering processing 20 trials to avoid the randomness perturbation and report the result with the lowest value for the objective function of k-means clustering among the 20 results.
The maximum number of iteration in all methods is set to 100.

\subsection{Comparison result and analysis}
The comparison results are shown in tables \ref{Tab:leaves_measures}, \ref{Tab:3sources_measures}, and \ref{Tab:hw_measures}. The best and second best value of each measure is shown in bold and underlined, respectively. We note that the standard deviation is avoided to be show for the k-means clustering trials since simply, not all the methods employ it, we mention that the values are rounded to the ten-thousandths.

\begin{table}[H]
\centering
\begin{minipage}{\textwidth} 
\caption{Clustering measure on HW data set}
\label{Tab:hw_measures}
\resizebox{\columnwidth}{!}{%
\begin{tabular}{l c c c c c c c c }
\hline
\textbf{Method} & \textbf{Fscore} & \textbf{Precision} & \textbf{Recall} & \textbf{NMI} & \textbf{ARI} & \textbf{ACC} & \textbf{Purity } \\ \hline
\textbf{AMGL} & 0,4011 & 0,3081 & 0,5788 & 0,6162 & 0,3115 & 0,5536 & 0,5893 \\ 
\textbf{MVGL} & 0.6367 & 0.5570 & 0.7429 & 0.7231 & 0.5900 & 0.6895 & 0.7245 \\ 
\textbf{Pairwise MLRSSC} & 0,7264 & \underline{0,8627} & 0,6276 & 0,7998 & 0,6907 & 0,747 & \underline{0,9185} \\ 
\textbf{Pairwise KMLRSSC} & 0,5399 & 0,6179 & 0,4794 & 0,5958 & 0,4818 & 0,6361 & 0,7412 \\ 
\textbf{Centroid KMLRSSC} & 0,5399 & 0,6179 & 0,4794 & 0,5958 & 0,4818 & 0,6361 & 0,7412 \\ 
\textbf{Centroid MLRSSC} & 0,7266 & \textbf{0,8678} & 0,6250 & 0,8021 & 0,6907 & 0,7438 & \textbf{0,9244} \\ 
\textbf{CGL} & \underline{0,8584} & 0,7941 & \underline{0,9630} & \underline{0,9354} & 0,8405 & \underline{0,8617} & 0,8741 \\ 
\textbf{CGMVC-NC} & \textbf{0.8996} & 0.8395 & \textbf{0.9853} & \textbf{0.9498} & \textbf{0.8874} & \textbf{0.8929} & 0.8996 \\ \hline
\end{tabular}
}
\end{minipage}
\end{table}

\begin{table}[H]
\centering
\begin{minipage}{\textwidth} 
\caption{Clustering measure on 3sources data set}
\label{Tab:3sources_measures}
\resizebox{\columnwidth}{!}{%
\begin{tabular}{l c c c c c c c c }
\hline
\textbf{Method} & \textbf{ Fscore } & \textbf{ Precision } & \textbf{ Recall } & \textbf{ NMI } & \textbf{ ARI } & \textbf{ ACC } & \textbf{ Purity } \\ \hline
\textbf{AMGL} & 0.2765 & 0.1711 & \textbf{0.7604} & \textbf{0.8490} & 0.2652 & 0.6352 & 0.7028  \\ 
\textbf{MVGL} & 0.3455 & 0.2255 & 0.7381 & 0.1522 & 0.0163 & 0.3846 & 0.4320 \\
\textbf{Pairwise MLRSSC} & \underline{0.6443} & 0.6045 & 0.6969 & 0.6167 & \underline{0.5460} & \underline{0.6840} & 0.7491  \\ 
\textbf{Pairwise KMLRSSC} & 0.4675 & 0.3798 & 0.6118 & 0.4837 & 0.3525 & 0.5098 & 0.5240 \\ 
\textbf{Centroid MLRSSC} & 0.5662 & 0.4629 & \underline{0.7492} & 0.6470 & 0.4725 & 0.6482 & 0.6683  \\ 
\textbf{Centroid KMLRSSC} & 0.4745 & 0.3847 & 0.6214 & 0.4824 & 0.3620 & 0.5169 & 0.5272  \\ 
\textbf{CGL} & 0.6224 & \underline{0.7097} & 0.5555 & \underline{0.6800} & 0.5263 & 0.6604 & \underline{0.7976}  \\ 
\textbf{CGMVC-NC} & \textbf{0.6670} & \textbf{0.7218} & 0.6213 & 0.6747 & \textbf{0.5760} & \textbf{0.7047} & \textbf{0.8133}  \\ 
\hline
\end{tabular}
}
\end{minipage}
\end{table}

\begin{table}[H]
\centering
\begin{minipage}{\textwidth} 
\caption{Clustering measure on 100Leaves data set}
\label{Tab:leaves_measures}
\resizebox{\columnwidth}{!}{%
\begin{tabular}{l c c c c c c c c }
\hline
\textbf{Method} & \textbf{Fscore} & \textbf{Precision} & \textbf{Recall} & \textbf{NMI} & \textbf{ARI} & \textbf{ACC} & \textbf{Purity} \\ \hline
\textbf{AMGL} & 0,2765 & 0,1711 & \underline{0,7604} & \underline{0,8490} & 0,2652 & 0,6352 & 0,7028 \\
\textbf{MVGL}	& 0.3455 	& 0.2255 	& 0.7381 &	 0.1522 	 & 0.0163 	& 0.3846 	& 0.4320 \\ 
\textbf{Pairwise MLRSSC} & 0,0102 & 0,6703 & 0,0245 & 0,3434 & 0,0297 & 0,0633 & 0,7702 \\ 
\textbf{Pairwise KMLRSSC} & 0,1120 & \textbf{0,8365} & 0,0600 & 0,5401 & 0,0962 & 0,0800 & \textbf{0,8911} \\ 
\textbf{Centroid MLRSSC} & 0,0492 & 0,6232 & 0,0256 & 0,3583 & 0,0317 & 0,0732 & 0,7374 \\ 
\textbf{Centroid KMLRSSC} & 0,0517 & 0,6128 & 0,0270 & 0,3634 & 0,0343 & 0,0736 & 0,7306 \\ 
\textbf{CGL} & 0,6245 & 0,7097 & 0,5654 & 0,6871 & \underline{0,5263} & \underline{0,6604} & 0,8024 \\
\textbf{CGMVC-NC} & \underline{0.7521} & \underline{0.6407} & \textbf{0.9235} & \textbf{0.9476} & \textbf{0.7493} &\textbf{ 0.7936} & \underline{0.8142} \\ 
\hline
\end{tabular}
}
\end{minipage}
\end{table}

\noindent
Our proposed approach performs well on the benchmark datasets. It outperforms the other methods on the accuracy evaluation metric, and does generally well, on the rest of the evaluation metrics. It performs better the CGL on almost all the metrics over the different dataset. 
\section{Conclusion}
In conclusion, the proposed method for multi-view clustering using a non-convex tensor norm has shown promising results in improving the clustering accuracy compared to traditional methods. By incorporating the tensor structure of multi-view data, the proposed method can effectively capture the underlying correlations between different views, leading to more accurate clustering results. Despite the non-convexity of the tensor norm, the proposed method can still be optimized efficiently using existing algorithms. Overall, CGMVC-NC provides a valuable tool for multi-view data analysis and has the potential to enhance our understanding of complex systems in various fields, including computer vision, neuroscience, and social network analysis. Further research can explore the application of this method to other types of data and the extension to other machine learning tasks.

\end{document}